\documentclass[10pt,twocolumn,letterpaper]{article}

\usepackage{cvpr}

\usepackage[accsupp]{axessibility}  
\usepackage{graphicx}
\usepackage{amsmath}
\usepackage{amssymb}
\usepackage{booktabs}
\usepackage{multirow}

\usepackage{flushend,cuted}

\usepackage[pagebackref,breaklinks,colorlinks]{hyperref}

\usepackage[capitalize]{cleveref}
\crefname{section}{Sec.}{Secs.}
\Crefname{section}{Section}{Sections}
\Crefname{table}{Table}{Tables}
\crefname{table}{Tab.}{Tabs.}
\newcommand\blfootnote[1]{%
\begingroup
\renewcommand\thefootnote{}\footnote{#1}%
\addtocounter{footnote}{-1}%
\endgroup
}

\begin{document}

\title{
CLIP is Also an Efficient Segmenter: A Text-Driven Approach for \\ 
Weakly Supervised Semantic Segmentation}

\author{
Yuqi Lin$^{1*}$~~~
Minghao Chen$^{1*\dagger}$~~~
Wenxiao Wang$^2$~~~
Boxi Wu$^2$~~~
Ke Li$^3$~~~ \\
Binbin Lin$^2$~~~
Haifeng Liu$^1$~~~
Xiaofei He$^1$~~~
\smallskip 
\\
$^1$State Key Lab of CAD\&CG, College of Computer Science, Zhejiang University
\\
$^2$School of Software Technology, Zhejiang University    $^3$Fullong Technology
\\
\tt\small \{linyq5566, minghaochen01\}@gmail.com
}
\maketitle
\blfootnote{*Equal contribution.} \\
\blfootnote{$^{\dagger}$Corresponding author.}

\begin{abstract}
\vspace{-1mm}
   Weakly supervised semantic segmentation (WSSS) with image-level labels is a challenging task. Mainstream approaches follow a multi-stage framework and suffer from high training costs. In this paper, we explore the potential of Contrastive Language-Image Pre-training models (CLIP) to localize different categories with only image-level labels and without further training.
   To efficiently generate high-quality segmentation masks from CLIP, we propose a novel WSSS framework called CLIP-ES. Our framework improves all three stages of WSSS with special designs for CLIP: 1) We introduce the softmax function into GradCAM and exploit the zero-shot ability of CLIP to suppress the confusion caused by non-target classes and backgrounds. Meanwhile, to take full advantage of CLIP, we re-explore text inputs under the WSSS setting and customize two text-driven strategies: sharpness-based prompt selection and synonym fusion. 2) To simplify the stage of CAM refinement, we propose a real-time class-aware attention-based affinity (CAA) module based on the inherent multi-head self-attention (MHSA) in CLIP-ViTs. 3) When training the final segmentation model with the masks generated by CLIP, we introduced a confidence-guided loss (CGL) focus on confident regions. 
   Our CLIP-ES achieves SOTA performance on Pascal VOC 2012 and MS COCO 2014 while only taking 10\% time of previous methods for the pseudo mask generation. Code is available at \href{https://github.com/linyq2117/CLIP-ES}{https://github.com/linyq2117/CLIP-ES}.

\end{abstract}

\begin{figure}[t]
  \centering
   \includegraphics[width=1.0\linewidth]{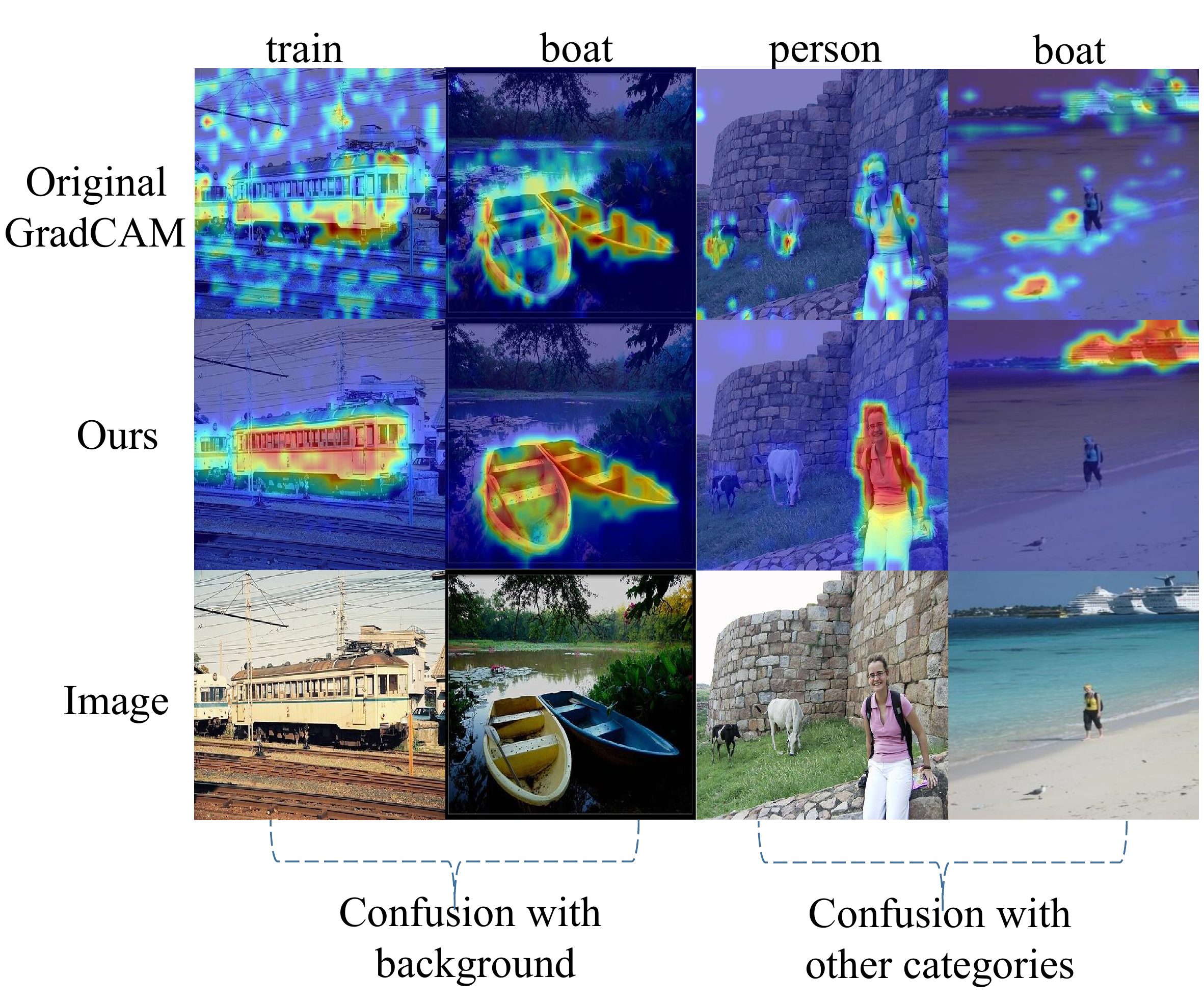}
    \vspace{-6mm}
   \caption{Effect of the softmax function on GradCAM of CLIP. The original GradCAM uses the logit (before the softmax) of the target class to compute gradient. We propose to compute gradient based on the probability (after the softmax). It can avoid confusion between the target class and background (the first two columns) and other object classes in the dataset (the last two columns).
   }
   \label{fig:1}
   \vspace{-2mm}
\end{figure}

\section{Introduction}
\label{sec:introduction}

Semantic segmentation~\cite{Strudel2021SegmenterTF,chen2017deeplab} aims to predict pixel-level labels but requires labor-intensive pixel-level annotations. Weakly supervised semantic segmentation (WSSS) is proposed to reduce the annotation cost. WSSS only requires weak supervision, \eg, image-level labels~\cite{Ahn2018PSA}, bounding boxes~\cite{Dai2015BoxSupEB,Papandreou2015WeaklyandSL}, points~\cite{bearman2016spoint} or scribbles~\cite{Lin2016ScribbleSupSC,Vernaza2017LearningRL}. The most commonly used one is WSSS with image-level annotations, which is the focus of our paper.

Previous WSSS approaches~\cite{Lee2021advcam, Wang2020SEAM, xu2022mctformer, Xie_2022_CLIMS} with image-level labels typically follow a three-stage framework. First, a classification model is trained on the specific dataset 
to generate initial CAMs (Class Activation Maps). Then, the initial CAMs are refined by the pixel affinity network~\cite{Ahn2018PSA,Ahn2019IRN} or extra saliency maps~\cite{Jiang2019OOA,Sun2020MCIS}. At last, the refined CAMs serve as the pseudo masks to train a semantic segmentation model. Obviously, this multi-stage framework is complicated as it needs to train multiple models at different stages, especially the separate classification model and affinity network in the first two stages. Although some end-to-end methods~\cite{Araslanov_2020_CVPR_single_stage,zhang2020reliability_single_stage} are proposed to improve efficiency, they tend to achieve poor performance compared to multi-stage methods. Therefore, it is a challenge to simplify the procedure of WSSS while maintaining its high performance.

Recently, the Contrastive Language-Image Pre-training (CLIP)~\cite{CLIP}, a model pre-trained on 400 million image-text pairs from the Internet to predict if an image and a text snippet are matched, has shown great success in the zero-shot classification. This dataset-agnostic model could transfer to unseen datasets directly. Besides, the powerful text-to-image generation ability of CLIP, \ie, DALL-E2~\cite{ramesh2022dalle2}, indicates the strong relation between texts and corresponding components in the image. On the other hand, multi-head self-attention (MHSA) in ViT~\cite{dosovitskiy2020vit} 
reflects semantic affinity among patches and has the potential to substitute for affinity network. Motivated by these, we believe CLIP with ViT architecture could simplify the procedure of WSSS and localize categories in the image through well-designed texts.

This paper proposes a new framework, CLIP-ES, to improve each stage in terms of efficiency and accuracy for WSSS.
In the first stage, the generated CAMs are usually redundant and incomplete. Most methods~\cite{Wang2020SEAM, Wu2021EmbeddedDA} are based on binary cross-entropy for multi-label classification. The loss is not mutually exclusive, so the generated CAMs suffer from confusion between foreground and non-target foreground categories, \eg, person and cow, or foreground and background categories, \eg, boat and water, as shown in \cref{fig:1}. The incompleteness stems from the gap between the classification and localization tasks, causing CAMs only focus on discriminative regions. To solve the confusion problems above, we introduce the softmax function into GradCAM to make categories mutually exclusive and define a background set to realize class-related background suppression. 
To get more complete CAMs and fully enjoy the merits inherited from CLIP, we investigate the effect of text inputs in the setting of WSSS and design two task-specific text-driven strategies: sharpness-based prompt selection and synonym fusion. 

In the second stage, instead of training an affinity network as in previous works, we leverage the attention obtained from the vision transformer. However, the attention map is class-agnostic, while the CAM is class-wise. To bridge this gap, we propose a class-aware attention-based affinity (CAA) module to refine the initial CAMs in real-time, which can be integrated into the first stage. Without fine-tuning CLIP on downstream datasets, our method retains CLIP’s generalization ability and is flexible to generate pseudo labels for new classes and new datasets.

In the last stage, the pseudo masks from the refined CAMs are viewed as ground truth to train a segmentation model in a fully supervised manner. However, the pseudo mask may be noisy and directly applied to training may mislead the optimization process. We proposed a confidence-guided loss (CGL) for training the final segmentation model by ignoring the noise in pseudo masks. 

Our contributions are summarized as follows:
\begin{itemize}
	\item We propose a simple yet effective framework for WSSS based on frozen CLIP. We reveal that given only image-level labels, CLIP can perform remarkable semantic segmentation without further training. Our method can induce this potential of localizing objects that exists in CLIP.
    \item We introduce the softmax function into GradCAM and design a class-related background set to overcome category confusion problems. To get better CAMs, some text-driven strategies inherited from CLIP are explored and specially redesigned for WSSS.
    \item We present a class-aware attention-based affinity module (CAA) to refine the initial CAMs in real time, and introduce confidence-guided loss (CGL) to mitigate the noise in pseudo masks when training the final segmentation model. 
    \item  Experiment results demonstrate that our framework can achieve SOTA performance and is 10x efficient than other methods when generating pseudo masks. 
\end{itemize}

\begin{figure*}
  \centering
  \includegraphics[width=0.99\linewidth]{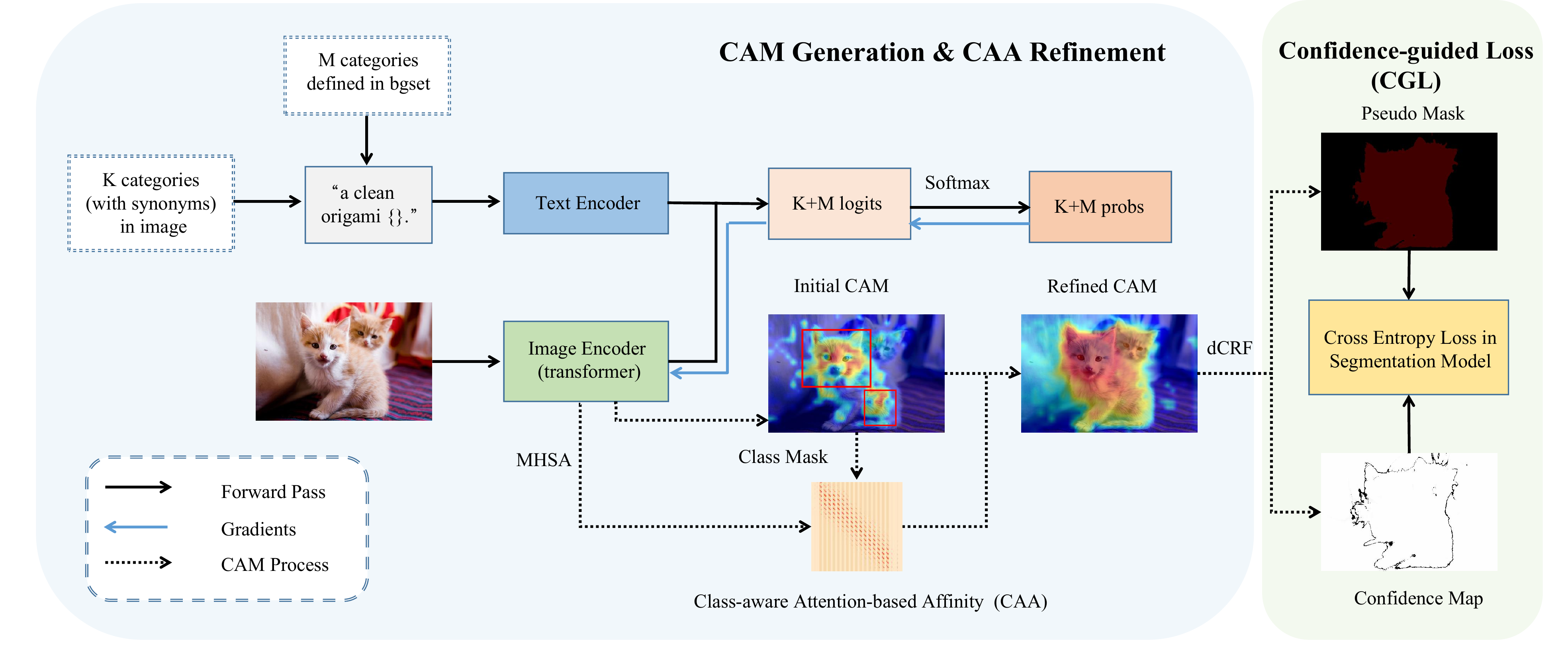}
  \vspace{-2mm}
  \caption{An overview of our proposed framework. We introduce the softmax function into GradCAM and define a class-related background set to make categories mutually exclusive. K and M represent the number of categories in an image and background set, respectively. The initial CAMs are generated by Grad-CAM with well-designed texts (\eg, prompt selection, synonym fusion). CAA module is proposed based on intrinsic MHSA in the transformer to refine the initial CAMs in real time. The whole CAM generation process is training-free. CGL ignores noisy positions when computing loss based on the confidence map.}
  \vspace{-2mm}
  \label{fig:framework overview}
\end{figure*}

\section{Related Work}
\label{sec:relatedwork}
\subsection{Weakly Supervised Semantic Segmentation}

Most existing approaches for WSSS train a classification network and extract localization maps from CNNs based on Class Activation Maps (CAMs)~\cite{cam}. However, the initial CAMs are usually incomplete or redundant. Several methods are proposed to improve the quality of CAMs and the final segmentation at different stages.

\textbf{Generating Initial CAM Stage.} In this stage, to address the incompleteness problem, some works train classification networks with auxiliary tasks, and additional losses are designed to guide the model to discover more object regions~\cite{Chang2020sc-cam,Wang2020SEAM,Ru2021LearningVW,Xu2021LeveragingAT}. ``Erasing'' is another strategy that erases an image's or feature map's discriminative parts to force the network to discover more regions~\cite{Kweon2021UnlockingTP,Wei2017ObjectRM,Hou2018SelfErasingNF}. Some works accumulate multiple activations in the training process~\cite{Jiang2019OOA,kim2021discriminative,Yao2021NSROM} and other works are from perspective of cross-image mining~\cite{Sun2020MCIS,Fan2020CIANCA,Li2021GroupWiseSM,Wu2021EmbeddedDA}, self-supervised mechanism~\cite{Chen_2022_CVPR_SIPE,Wang2020SEAM} and anti-adversarial attack~\cite{Lee2021advcam}. To solve the redundancy problem, previous works use softmax cross entropy as an additional loss to reactivate the model~\cite{recam} or introduce extra out-of-distribution(OoD) data~\cite{lee2022w--ood}. Recently, some transformer-based methods~\cite{xu2022mctformer, AFA} appear in the WSSS task and achieve competitive performance.

\textbf{Refining Initial CAM Stage.} In this stage, pairwise semantic affinity is typically learned to refine CAM maps. PSA~\cite{Ahn2018PSA} trains a network to learn pixel affinity and propagate the semantics of strong responses in attention maps to semantically similar pixels. IRNet~\cite{Ahn2019IRN} and BES~\cite{Chen2020bes} synthesize class boundaries and expand the object coverage until boundaries. Another approach exploits additional saliency maps to obtain precise background or distinguish co-occurring objects~\cite{Lee2021EPS,Fan2020ICD,jiang2022l2g}.

\textbf{Training Segmentation Model Stage.} Traditional methods~\cite{Lee2021EPS,jiang2022l2g} generate pseudo masks from CAMs by applying a global threshold, which can't fully utilize CAMs due to ignorance of confidence information. 
Only a few works attempt to suppress the noise at this stage. PMM~\cite{li2021pseudo_matters} proposes the pretended under-fitting strategy to reweight losses of potential noise pixels. URN~\cite{Li2022URN} scales the prediction map multiple times for uncertainty estimation. However, the former is merely operated on the loss level and doesn't use confidence while the latter is time-consuming for multiple dense CRF processes.

\subsection{Contrastive Language-Image Pretraining}

Contrastive Language-Image Pretraining (CLIP)~\cite{CLIP} consists of an image encoder and a text encoder. It learns corresponding embeddings and measures the similarity between images and texts. Benefiting from this flexible framework, CLIP can be trained on super-large datasets and is widely used on the downstream zero-shot task.
CLIMS~\cite{Xie_2022_CLIMS} first introduced CLIP into WSSS to activate more complete object regions and suppress background regions. However, in CLIMS, CLIP is just a tool to evaluate the existence of objects and another CNN model is used to generate CAMs. In this paper, we directly use CLIP to generate CAMs and thoroughly explore the relationship between the text and objects in the image, which is more simple and more efficient.

\section{Method}
\label{sec:method}
In this section, we propose our CLIP-ES framework, which is depicted in \cref{fig:framework overview}. We first review GradCAM and CLIP, and demonstrate the effect of the softmax function on GradCAM with the corresponding class-related background suppression strategy. 
Then, we introduce two text-driven strategies proposed for CLIP in the WSSS setting: sharpness-based prompt selection and synonym fusion. Finally, we present class-aware attention-based affinity (CAA) and confidence-guided loss (CGL) in detail.

\subsection{Softmax-GradCAM}
\label{sec3.1}
Class Activation Mapping (CAM)~\cite{cam} is widely used to identify the discriminative regions for the target class by the weighted combination of feature maps. However, it is only applicable to specific CNN architectures, \eg, models with a global average pooling (GAP) layer immediately after the feature maps. 
GradCAM~\cite{gradcam} uses the gradient information to combine feature maps and thus there is no requirement for network architecture. For original GradCAM, the class feature weights can be calculated as \cref{eq1}:
\begin{equation}
  w_k^c = \frac{1}{Z}\sum_i\sum_j \frac{\partial Y^c}{\partial A_{ij}^k}
  \label{eq1}
\end{equation}
where $w_k^c$ is the weight corresponding to \textit{c-th} class for \textit{k-th} feature map, $Z$ is the number of pixels in the feature map, $Y^c$ is the logit score for \textit{c-th} class and $A_{ij}^k$ represents the activation value for \textit{k-th} feature map at location $(i,j)$. Then the CAM map of class \textit{c} at spatial location $(i,j)$ can be obtained by \cref{eq2}. \textit{ReLU} is adopted to ignore features that negatively influence the target class.
\begin{equation}
  CAM_{ij}^c = ReLU\bigg (\sum_k w_k^cA_{ij}^k\bigg )
  \label{eq2}
\end{equation}

Pretrained CLIP models include two architectures, \eg, ResNet-based and ViT-based. Note that Grad-CAM is not only applicable to CNN-based architecture but also works on the vision transformer. In this paper, we leverage the ViT-based CLIP model because the CNN-based model fails to explore the global context and suffers from the discriminative part domain heavily. The comparison between these two architectures can be found in Appendix.

Our work adapts GradCAM to CLIP. In vanilla GradCAM~\cite{gradcam}, the final score is the logits before the softmax function. Due to the multi-label setting of WSSS, the classification network often employs the binary cross entropy loss~\cite{Wang2020SEAM,Wu2021EmbeddedDA}, thus lacking competition among different classes. 
CLIP is trained by cross-entropy loss with softmax, but it still suffers from the category confusion problem in our experiment. We assume it is because the training data of CLIP are image-text pairs rather than a fixed set of separate categories. For an image, the corresponding text snippet could contain visual concepts of several classes, which can't compete with each other through softmax either. 
This paper introduces the softmax function into GradCAM to make different categories mutually exclusive.
Specifically, the final score is computed by softmax as follows:
\begin{equation}
  s^c = \frac{\exp(Y^c)}{\sum_{c'=1}^C \exp(Y^{c'})}
  \label{eq3}
\end{equation}
$s^c$ is the score for \textit{c-th} class after softmax. The processed scores are then used to compute the gradient, and the class feature weights can be calculated as:
\begin{equation}
\begin{aligned}
    w_k^c &= \frac{1}{Z}\sum_i\sum_j\sum_{c'} \frac{\partial Y^{c'}}{\partial A_{ij}^k} * \frac{\partial s^c}{\partial Y^{c'}} \\
    &= \frac{1}{Z}\sum_i\sum_j \frac{\partial Y^c}{\partial A_{ij}^k} * s^c(1-s^c) \\
    &+ \frac{1}{Z}\sum_i\sum_j\sum_{c'\neq c} \frac{\partial Y^{c'}}{\partial A_{ij}^k} * s^c(-s^{c'})
  \label{eq4}
\end{aligned}
\end{equation}

\cref{eq4} indicates that the weight of the target feature map will be suppressed by non-target classes. So the corresponding CAMs of the target class can be revised by the remaining classes. However, the competition is only limited to categories defined in the dataset. To disentangle pixels of the target class from background classes, 
we propose a class-related background suppression method. We define a background category set containing $M$ common class-related categories for classes defined in datasets. 
In this way, pixels of background categories will be suppressed. Thanks to the zero-shot capability of CLIP, we only need to revise input texts rather than retrain the classification network for background categories like previous training-based methods.

\begin{figure}[t]
  \centering
   \includegraphics[width=0.8\linewidth]{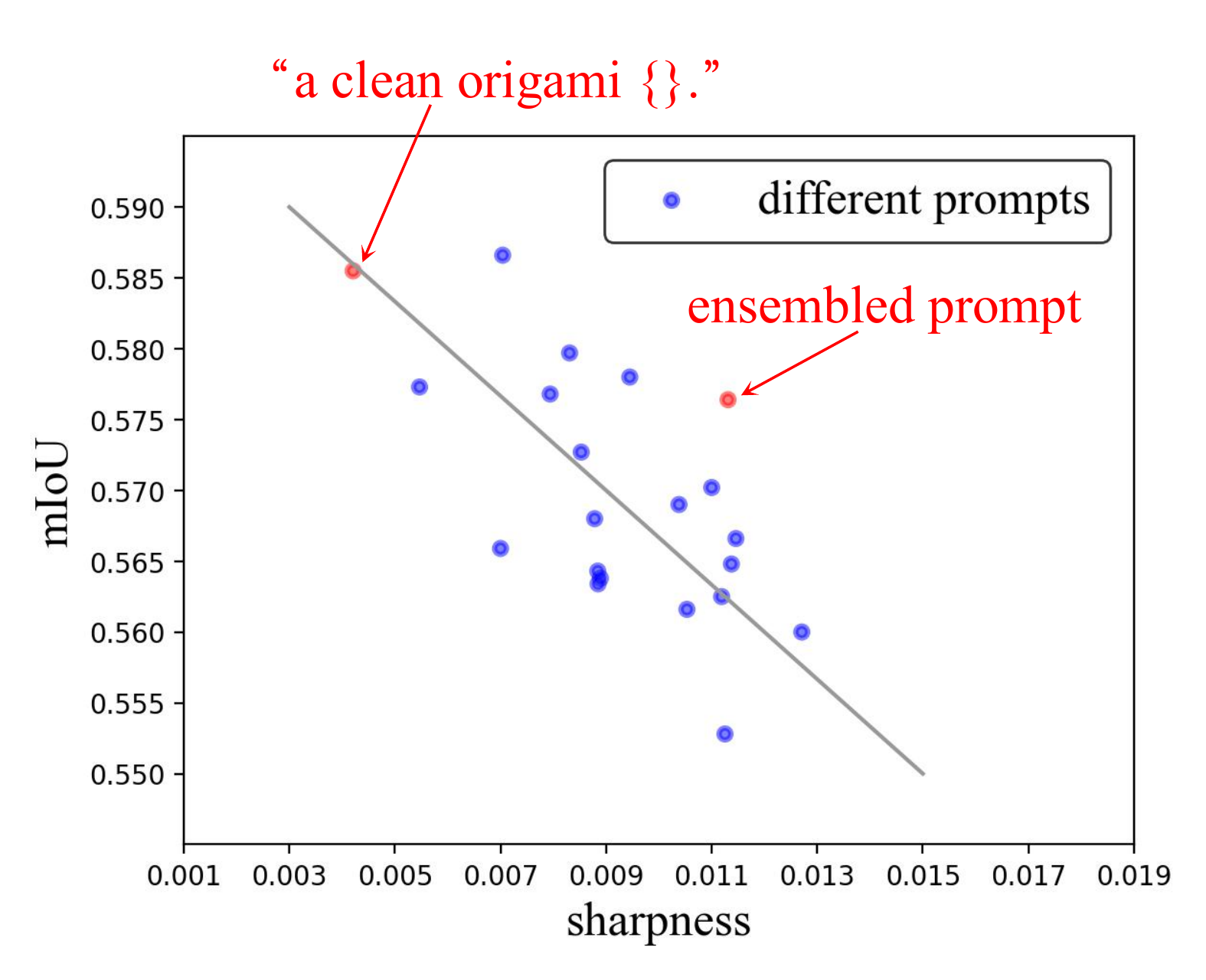}
    \vspace{-2mm}
   \caption{Relations between $sharpness$ and $mIoU$ using different prompts on PASCAL VOC 2012 train set. ``a clean origami \{\}.'' is the prompt we finally adopted in our paper. 
   }
   \vspace{-2mm}
   \label{fig:sharpness vs miou}
\end{figure}

\subsection{Text-driven Strategies}
For CLIP, the text encoder acts as a linear classifier weight generator 
based on the text specifying the visual concepts the classes represent. Our framework could enjoy multiple merits inherited from CLIP by designing specific text inputs.  
In this part, we re-explore the effect of text inputs under the WSSS setting and propose sharpness-based prompt selection and synonyms fusion to boost the CAM generation process.
\vspace{-2mm}
\subsubsection{Sharpness-based Prompt Selection}
\ 
\indent We find that the performance of prompt ensembling differs between the classification task and the WSSS task. Specifically, prompt ensembling can outperform every single prompt by a large margin for the classification task on ImageNet~\cite{deng2009imagenet}, while it is not the optimal choice when performing WSSS on PASCAL VOC~\cite{everingham2010pascal}.
We suspect this difference is primarily due to the varying amount of labels per image.
The classification dataset, \eg, ImageNet, is single-labeled, while the segmentation dataset, \eg, PASCAL VOC, is multi-labeled. The former aims to assign a maximum score for the unique target class, while the latter need to consider all target classes in an image. We claim that prompt ensembling will make the target class with the top score more prominent. 
But for multi-labeled images, a prominent target class will suppress scores of other target classes. This affects subsequent gradient computing for GradCAM and leads to poor segmentation performance. 

To verify our conjecture, we design a metric, namely \textit{sharpness}, to measure the distribution of target class scores for multi-label images using different prompts. This metric is inspired by \textit{Coefficient of Variation}, a metric widely used in statistics. Assume there are $n$ images in the dataset and $k(k>=1)$ classes in an image, the \textit{sharpness} based on a specific prompt can be calculated as follows:
 \begin{equation}
  sharpness(\operatorname{prompt}) = \frac{\sum_i^n var(s_{i1},...,s_{ik})}{\sum_i^n mean(s_{i1},...,s_{ik})}
  \label{eq5}
\end{equation}
$s_{ij}$ represents scores for \textit{j-th} class after softmax in \textit{i-th} image. Since \textit{Coefficient of Variation} is unstable when \textit{mean} is close to 0, we use variance instead of standard deviation to highlight the effect of dispersion. 

In \cref{fig:sharpness vs miou}, we compare \textit{sharpness} and corresponding segmentation results among 20 prompts randomly selected from the ImageNet prompts used in CLIP\footnote{\href{https://github.com/openai/CLIP/blob/main/notebooks/Prompt_Engineering_for_ImageNet.ipynb}{https://github.com/openai/CLIP/blob/main/notebooks}} on Pascal VOC 2012 train set. As the result demonstrates, our proposed metric is roughly negatively correlated to segmentation performance. 
Consequently, \textbf{\textit{sharpness} can serve as convenient guidance for prompt choice, and only image-level labels are needed.} After trial and error, we find that some abstract descriptions, \eg, \textit{``origami''} and \textit{``rendering''}, and some adjectives, \eg, \textit{``clean''}, \textit{``large''} and \textit{``weird''}, have a positive effect on segmentation performance. We finally select \textit{``a clean origami \{\}.''} as our prompt, which has the lowest \textit{sharpness}. 
\vspace{-2mm}
\subsubsection{Synonym Fusion} 
\ 
\indent Since the category names provided in the datasets are limited, we use synonyms to enrich semantics and disambiguate. There are various strategies to merge semantics of different synonyms, \eg, sentence-level, feature-level, or CAM-level. We provide a detailed comparison of the three strategies in the Appendix. In this paper, we merge synonyms at the sentence level. Specially, we put different synonyms into one sentence, \eg, ``A clean origami of person, people, human''. This can disambiguate when facing polysemous words and is time-efficient as other methods require multiple forward passes. The synonyms can easily be obtained from WordNet or the nearest Glove word embedding. In addition, the performance of some classes can be further improved by customizing specific words. For example, CAMs of ``person'' tend to focus on faces, while the ground truth segmentation masks cover the whole body. It is likely that ``person'' and ``clothes'' are treated as two different categories in CLIP. By replacing ``person'' with ``person with clothes'', this problem can be alleviated to some extent. 

\subsection{Class-aware Attention-based Affinity (CAA)}
Recently, some works~\cite{xu2022mctformer, AFA} use attention obtained from the transformer as semantic-level affinity to refine initial CAMs. But the improvement is limited and they still require an additional network~\cite{xu2022mctformer} or extra layers~\cite{AFA} to further refine CAMs. 
It is because the original multi-head self-attention (MHSA) is class-agnostic, while the CAM is class-wise. Leveraging MHSA directly may amplify noise by propagating noisy pixels to its semantically similar regions during refinement, as is shown in \cref{fig:ablation_caa}. 

We propose class-aware attention-based affinity (CAA) to improve vanilla MHSA. Given an image, we can get the class-wise CAM map $M_c \in R^{h \times w}$ for each target class $c$ and the attention weight $W^{attn} \in R^{hw \times hw}$ from MHSA. For the attention weight, which is asymmetric because of the different projection layers used by the query and key, we leverage Sinkhorn normalization~\cite{Sinkhorn1964ARB} (alternately applying row-normalization and column-normalization) to convert it to a doubly stochastic matrix $D$, and the symmetric affinity matrix $A$ can be obtained as follows:
\begin{equation}
  A = \frac{D + D^T}{2},  where D = Sinkhorn(W^{attn}).
  \label{eq6}
\end{equation}
For the CAM map $M_c \in R^{h \times w}$, we can obtain a mask map for each target class $c$ by thresholding the CAM of this class with $\lambda$. We find connected regions on the mask map and use the minimum rectangle bounding boxes covering those connected regions. These boxes mask the affinity weight $A$, and then each pixel can be refined based on the masked affinity weight by its semantically similar pixels. We employ the bounding box mask rather than the pixel mask to cover more regions of the objects for the extreme incompleteness of initial CAMs. We repeat this refinement multiple times, and this process can be formalized as follows.
\begin{equation}
  M_c^{aff} = B_c \odot A^t \cdot vec(M_c)
  \label{eq7}
\end{equation}
where $B_c \in R^{1 \times hw}$ is box mask obtained from CAM of class $c$, $\odot$ is Hadamard product, $t$ denotes the number of refining iterations and $vec(\cdot)$ means vectorization of a matrix. Note that we extract the attention map and CAM with the same forward pass. Hence, CAA refinement is real-time and requires no additional stage like previous works. 


\subsection{Confidence-guided Loss (CGL)}
Each pixel in the CAM indicates the confidence of this position belonging to the target class. Most methods generate pseudo masks from CAMs by simply setting a threshold to distinguish target objects and backgrounds. It may bring noise into pseudo masks because those positions with low confidence are too uncertain to belong to the correct class. Thus, we attempt to ignore those unconfident positions and propose a confidence-guided loss (CGL) to make full use of generated CAMs. Specifically, given CAM maps $X \in R^{h \times w \times c}$ of an image with $c$ target classes, the confidence map can be obtained as:
\begin{equation}
  Conf(i,j) = \max (1-\max_c(X(i,j,c)), \max_c(X(i,j,c)))
  \label{eq8}
\end{equation}
and the final loss is defined as \cref{eq9}:
\begin{equation}
  \hat L(i,j) = 
  \begin{cases}
  L(i,j),\quad & Conf(i,j)>=\mu \\
  0, \quad & Conf(i,j)<\mu
  \end{cases}
  \label{eq9}
\end{equation}
where $L(i,j)$ is the cross entropy loss between the prediction of the semantic segmentation model and the pseudo mask for pixel $(i,j)$, and $\mu$ is a hyper-parameter to ignore pixels with low confidence.


\section{Experiments}
\label{sec:experiments}
\subsection{Experimental Setup}

\textbf{Datasets and Evaluation Metric.}
We evaluate our proposed framework on PASCAL VOC 2012~\cite{everingham2010pascal} and MS COCO 2014~\cite{lin2014microsoftcoco} datasets. PASCAL VOC 2012 contains 21 categories (one background category). An augmented set with 10,582 images is used for training following~\cite{Lee2021EPS, Lee2021advcam}. MS COCO 2014 contains 80 object classes and one background class. It includes 82,081 images for training and 40,137 images for validation. We only used image-level ground-truth labels during CAM generation. The mean Intersection over Union (mIoU) is adopted as the evaluation metric for all experiments.

\textbf{Implementation Details.} For CAM generation, we adopt CLIP pre-trained model ViT-B-16~\cite{CLIP}. The feature map used to generate CAM is the one before the last self-attention layer in ViT. 
We replace the class token with the average of remaining tokens to compute final logits, which can significantly boost the performance. Detailed analysis is discussed in Appendix. Input images remain their original size, and we do not use the multi-scale strategy during inference. 
$\lambda$ used in the CAA module is set to 0.4 and 0.7 for VOC and COCO, respectively. The generated CAMs are further post-processed by dense CRF\cite{CRF} to generate final pseudo masks. For final segmentation, we use ResNet101-based DeepLabV2 following prior works ~\cite{Lee2021advcam,Lee2021EPS,Xie_2022_CLIMS}, and more details are provided in Appendix.

\begin{table}
  \centering
  \begin{tabular}{lccc}
    \toprule
    Method  & Seed   &  dCRF & RW\\
    \midrule
    IRN~\cite{Ahn2019IRN}       & 48.8        & 54.3  &  66.3\\
    SC-CAM~\cite{Chang2020sc-cam}  & 50.9        & 55.3  & 63.4\\
    SEAM~\cite{Wang2020SEAM}  & 55.4       & 56.8  &  63.6\\
    AdvCAM~\cite{Lee2021advcam}  & 55.6       & 62.1 & 68.0 \\
    CLIMS~\cite{Xie_2022_CLIMS} & 56.6 & 62.4 & 70.5 \\
    RIB~\cite{Lee2021ReducingIB} & 56.5 & 62.9 & 70.6 \\
    OoD~\cite{lee2022w--ood} & 59.1  &  65.5 & 72.1 \\
    MCTfomer~\cite{xu2022mctformer} & 61.7 & 64.5 & 69.1 \\
    Ours          & \textbf{70.8}   &  \textbf{75.0} & - \\
    \bottomrule
  \end{tabular}
  \caption{mIoU of generated CAMs on PASCAL VOC 2012 train set. dCRF denotes using dense CRF~\cite{CRF} to post-process CAMs. RW represents training affinity networks to refine CAMs.}
  \label{tab:cam_quality}
\end{table}

\begin{table}
  \centering
  \begin{tabular}{lcccc}
    \toprule
    Method      &     mIoU  \\
    \midrule
    Initial     &    58.6 / 62.4$^*$  \\
    \midrule
    Initial + MHSA &   68.2 / 67.0$^*$ \\
    Initial + CAA &   70.8 / 70.5$^*$ \\
    \midrule
    Initial + MHSA + dCRF & 72.1 / 70.1$^*$ \\
    Initial + CAA + dCRF & \textbf{75.0} / 74.1$^*$ \\
    \bottomrule
  \end{tabular}
  \caption{mIoU of initial CAMs, CAA refined CAMs, and vanilla MHSA refined CAMs on PASCAL VOC 2012 train set. $^*$ means adopting the multi-scale strategy during inference.}
  \vspace{-4mm}
  \label{tab:caa}
\end{table}


\begin{table*}[]
\centering
\begin{tabular}{lcccccc}
\toprule
\multirow{2}{*}{Method} & \multicolumn{2}{c}{Classification Time}                               & \multirow{2}{*}{dCRF} & \multirow{2}{*}{Affinity} & \multirow{2}{*}{Total Time} & \multirow{2}{*}{Memory Cost} \\ 
& \multicolumn{1}{c}{Train} & \multicolumn{1}{c}{Inference} & &  &  &  \\ 
\midrule
AdvCAM~\cite{Lee2021advcam}  & \multicolumn{1}{c}{-}  & \multicolumn{1}{c}{70.5}      & 0.2 & 6.5  & 77.2  & 18G     \\ 
CLIMS~\cite{Xie_2022_CLIMS}  & \multicolumn{1}{c}{2.1}   & \multicolumn{1}{c}{\ \ 0.3}       &   0.2  & 6.5  & \ \ 9.1  &   18G\\ 
MCTformer~\cite{xu2022mctformer}  & \multicolumn{1}{c}{0.5}   & \multicolumn{1}{c}{\ \ 2.5}  & -  & 3.0  & \ \ 6.0  &   18G  \\ 
Ours  & \multicolumn{1}{c}{-}     & \multicolumn{1}{c}{\ \ 0.4} & 0.2  & -  & \textbf{\ \ 0.6}& \textbf{\ \ 2G} \\ \bottomrule
\end{tabular}
\caption{Time and memory cost of different methods to generate pseudo masks on PASCAL VOC train aug set (containing 10582 images in total). The time unit is \textbf{hour} and the memory unit is \textbf{GB}. Note that the inference and dCRF processes are combined in MCTformer. }
  \label{tab:time cost}
\end{table*}

\begin{table}
  \centering
  \begin{tabular}{lclcc}
    \toprule
    Method  & Backbone   & Seg.    & Val   & Test \\
    \midrule
    \multicolumn{5}{l}{\textbf{Image-level supervision + Saliency maps.}} \\
    
    OAA+~\cite{Jiang2019OOA}      & R101       & V1$^\ddagger$    & 65.2  & 66.4\\
    MCIS~\cite{Sun2020MCIS}      & R101       & V1$^\ddagger$    & 66.2  & 66.9 \\
    ICD~\cite{Fan2020ICD}       & R101       & V1$^\ddagger$    & 67.8  & 68.0 \\
    NSROM~\cite{Yao2021NSROM}     & R101       & V2$^\ddagger$    & 70.4  & 70.2 \\
    DRS~\cite{kim2021discriminative}       & R101       & V2$^\ddagger$    & 71.2  & 71.4 \\
    EPS~\cite{Lee2021EPS}       & R101       & V2$^\ddagger$    & 70.9  & 70.8 \\
    EDAM~\cite{Wu2021EmbeddedDA}      & R101       & V1$^\ddagger$    & 70.9  & 70.6 \\
    RIB~\cite{Lee2021ReducingIB}       & R101       & V2               & 70.2  & 70.0 \\
    L2G~\cite{jiang2022l2g}       & R101       & V2$^\ddagger$    & 72.1  & 71.7 \\
    RCA~\cite{zhou2022regional}       & R101       & V2$^\ddagger$    & 72.2  & 72.8 \\
    PPC+EPS~\cite{du2022weakly}   & R101       & V2               & 72.6  & 73.6 \\
    \midrule
    \multicolumn{5}{l}{\textbf{Image-level supervision only.}} \\
    
    PSA~\cite{Ahn2018PSA}       & WR38       & V1               & 61.7  & 63.7 \\
    IRN~\cite{Ahn2019IRN}       & R50        & V2               & 63.5  & 64.8 \\
    ICD~\cite{Fan2020ICD}       & R101       & V1$^\ddagger$    & 64.1  & 64.3 \\
    SEAM~\cite{Wang2020SEAM}      & WR38       & V1               & 64.5  & 65.7 \\
    SC-CAM~\cite{Chang2020sc-cam}    & R101       & V2$^\ddagger$    & 66.1  & 65.9 \\
    BES~\cite{Chen2020bes}       & R101       & V2$^\ddagger$    & 65.7  & 66.6 \\
    AdvCAM~\cite{Lee2021advcam}    & R101       & V2               & 68.1  & 68.0 \\
    SIPE~\cite{Chen_2022_CVPR_SIPE}      & R101       & V2$^\ddagger$    & 68.8  & 69.7 \\
    RIB~\cite{Lee2021ReducingIB}       & R101       & V2               & 68.3  & 68.6 \\
    ReCAM~\cite{recam}     & R101       & V2               & 68.5  & 68.4 \\
    AMN~\cite{Lee2022AMN}       & R101       & V2$^\ddagger$    & 70.7  & 70.6 \\
    MCTformer~\cite{xu2022mctformer} & WR38       & V1$^\dagger$     & 71.9  & 71.6 \\
    \midrule
    \multicolumn{5}{l}{\textbf{Image-level supervision + Language supervision.}} \\
    
    CLIMS~\cite{Xie_2022_CLIMS}     & R101       & V2               & 69.3  & 68.7 \\
    CLIMS~\cite{Xie_2022_CLIMS}     & R101       & V2$^\ddagger$    & 70.4  & 70.0 \\
    Ours   & R101  & V2               & \textbf{71.1} & \textbf{71.4} \\
    Ours   & R101  & V2$^\ddagger$    & \textbf{73.8} & \textbf{73.9} \\
    \bottomrule
  \end{tabular}
  \caption{Evaluation results on PASCAL VOC 2012 validation and test sets. The best results are in \textbf{bold}. Seg. denotes segmentation network. $^\dagger$ and $^\ddagger$ represents adopting VOC and MS COCO pretrained model, respectively.}
  \vspace{-4mm}
  \label{tab:segmentation performance VOC}
\end{table}

\subsection{Experimental Results}

\begin{figure*}[t]
  \centering
   \includegraphics[width=0.85\linewidth]{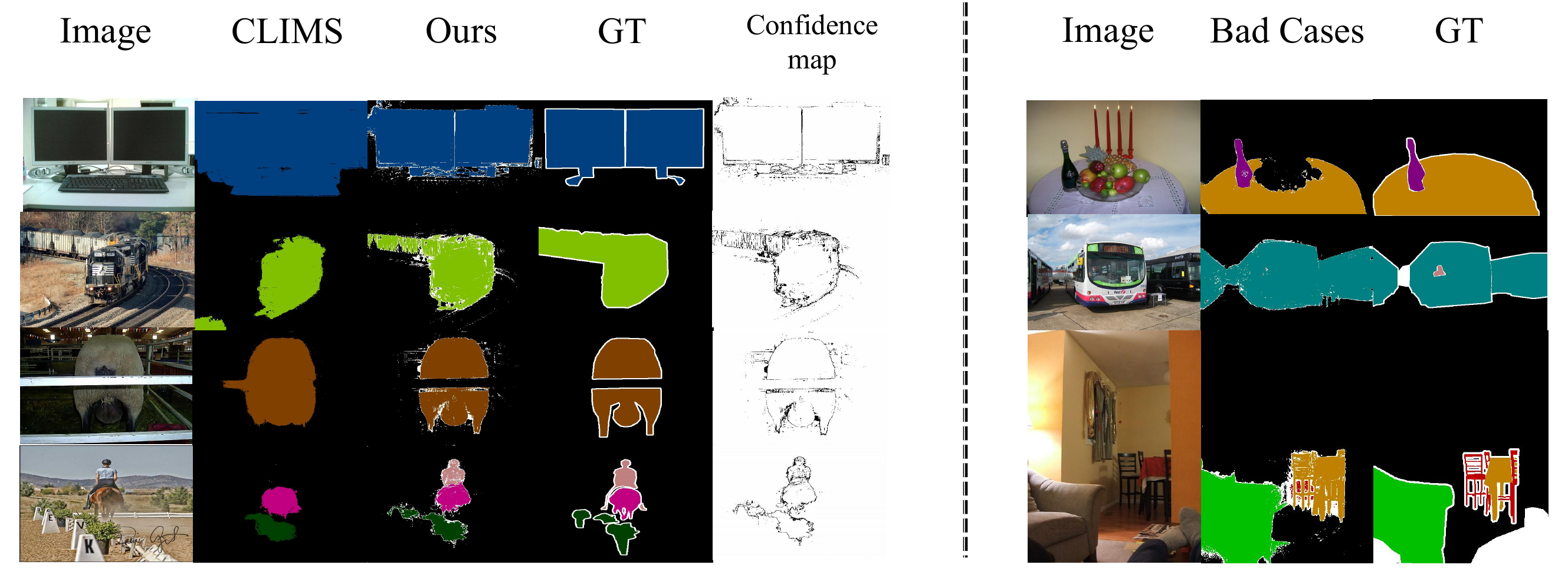}

   \caption{\textbf{Left:} Visualization of the pseudo masks generated by our framework and CLIMS. \textbf{Right:} Visualization of some bad cases.
   }
   \label{fig:vis_pseudo_label}
\end{figure*}

\textbf{Quality of Generated CAMs.} \cref{tab:cam_quality} shows the quality of our generated CAMs. 
Our framework outperforms all previous methods by a large margin on initial seeds. CRF could 
further boosts the performance to 75.0\%, which even outperforms previous methods with extra affinity networks. The result is accurate enough, hence the stage of training an affinity network is omitted. We show qualitative results of our framework and another language-guided method CLIMS~\cite{Xie_2022_CLIMS} in \cref{fig:vis_pseudo_label}. Our framework can produce accurate and complete segmentation masks. The bad cases mainly stem from occlusion and small objects, which are challenging even in a fully supervised setting. In addition, it is a common practice to aggregate the prediction results from multi-scale images during inference in previous works. In \cref{tab:caa}, we compare CAM quality generated by single-scale and multi-scale strategies (denoted with $^*$). The multi-scale inference has no improvement with the CAA module and dense CRF postprocessing and thus single-scale inference is adopted in our experiments.

\textbf{Time and Memory Efficiency.} In \cref{tab:time cost}, we compare our time and memory costs with some related works. Benefiting from the pre-trained CLIP model, our method requires no classification training on specific datasets. The CAA module intrinsic in ViT is integrated into the first stage that generates initial CAMs. Thus, our framework can refine CAMs in real time and requires no additional refinement stage by training affinity networks, \eg, PSA~\cite{Ahn2018PSA} and IRN~\cite{Ahn2019IRN}. The maximum memory occurs during the affinity network training for previous works, which is about 18GB for both PSA and IRN. As a result, our method is more than 10x efficient than other works in terms of time and memory. Meanwhile, inference speed is ensured by adopting the single-scale strategy, which is competitive with the multi-scale strategy in our approach (Tab.~\ref{tab:caa}). 

\textbf{Segmentation Performance.} To further evaluate the quality of pseudo masks, we train the segmentation model based on DeepLabV2 with ResNet-101 following~\cite{Chen2020bes,Chen_2022_CVPR_SIPE,Xie_2022_CLIMS}. In \cref{tab:segmentation performance VOC}, we compare our framework with related methods on PASCAL VOC 2012. Our method outperforms all previous works, even those with saliency maps as auxiliary supervision. Our CLIP-ES achieves 73.8\% and 73.9\% mIoU on the validation and test set, respectively, which is a new state-of-the-art. The evaluation results on MS COCO 2014 are reported in \cref{tab:segmentation performance COCO}. Our method also achieves the best performance, with 45.4\% mIoU on the validation set.

\begin{table}
  \centering
  \begin{tabular}{lcclc}
    \toprule
    Method  & Backbone   & Seg.    & Sup.    & Val \\
    \midrule
    EPS~\cite{Lee2021EPS}       & VGG16      & V2    & I+S  & 35.7 \\
    L2G~\cite{jiang2022l2g}       & R101       & V2    & I+S  & 44.2 \\
    IRN~\cite{Ahn2019IRN}       & R50        & V2    & I    & 32.6 \\
    IRN~\cite{Ahn2019IRN}       & R101       & V2    & I    & 41.4 \\
    URN~\cite{Li2022URN}       & R101       & PSPnet & I    & 40.7 \\
    SIPE~\cite{Chen_2022_CVPR_SIPE}      & R101       & V2    & I    & 40.6 \\
    RIB~\cite{Lee2021ReducingIB}       & R101       & V2    & I    & 43.8 \\
    AMN~\cite{Lee2022AMN}       & R101       & V2    & I    & 44.7 \\
    Ours       & R101       & V2    & I+L    & \textbf{45.4} \\
    \bottomrule
  \end{tabular}
  \caption{Evaluation results on MS COCO 2014 validation set. The best results are shown in \textbf{bold}. Seg. denotes segmentation network, and Sup. denotes the weak supervision type.}
  \label{tab:segmentation performance COCO}
  \vspace{-2mm}
\end{table}


\subsection{Ablation Study}

\begin{table}
  \centering
  \begin{tabular}{lccc}
    \toprule
    Method  & total & boat & train \\
    \midrule
    w/o softmax  & 49.4$^*$ / 49.4 & 24.1 & 43.8 \\
    with softmax & 53.3$^*$ / 58.6 & 46.9 & 57.5 \\
    \bottomrule
  \end{tabular}
  \caption{Ablation study of softmax function on VOC train set. $^*$ denotes only 20 categories defined in the dataset are used. Results are based on the initial CAMs and not refined by the CAA module.}
  \vspace{-4mm}
  \label{tab:softmax ablation}
\end{table}

\begin{figure}[t]
  \centering
   \includegraphics[width=0.85\linewidth]{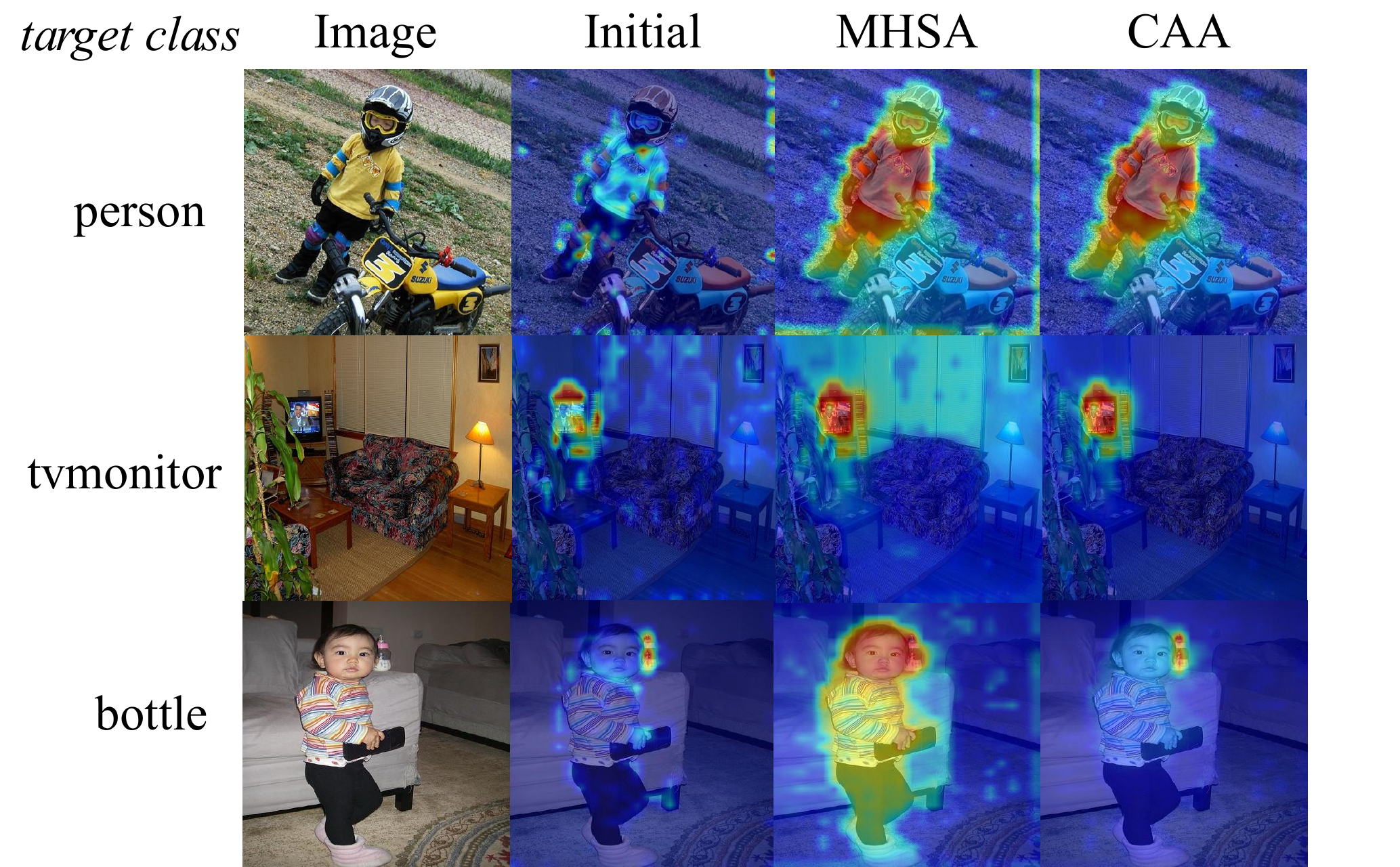}

   \caption{The initial CAMs generated by our proposed framework and comparison between CAA and MHSA refinement.
   }
   \vspace{-2mm}
   \label{fig:ablation_caa}
\end{figure}

\textbf{Effect of Softmax Function.} We introduce the softmax function into GradCAM to make categories mutually exclusive. First, 20 classes defined in VOC with and without softmax are compared. Results in \cref{tab:softmax ablation} (denoted with $^*$) show that softmax-based GradCAM can boost the performance remarkably (from 49.4\% to 53.3\%). Afterwards, to evaluate the effectiveness of the class-related background set we defined, we report results of ``boat'' (usually confused with ``water'') and ``train'' (usually confused with ``railway'') following \cite{Xie_2022_CLIMS,Lee2021EPS}. As \cref{tab:softmax ablation} shown, $mIoU$ can be improved by 22.8\% and 13.7\% for boat and train, respectively. The overall performance improves by 9.2\% among all classes. The results above suggest that softmax could solve the categories confusion problem efficiently.

\textbf{Effect of CAA.} In \cref{tab:caa}, we provide mIoU of the initial and CAA refined CAMs and compare our CAA module with vanilla MHSA in ViT. Results demonstrate that our CAA module can improve MHSA remarkably by introducing the class-aware mask. 
\cref{fig:ablation_caa} shows the visual comparison of different refinement strategies. Our CAA module could make object activations of the initial CAMs complete and mitigate the effect of falsely activated regions.

\textbf{Effect of CGL.} 
In \cref{tab:cgl ablation}, we compare CGL with the original Cross Entropy Loss. Results show that CGL can further boost performance. Note that it requires no additional information and merely fully utilizes the confidential information in CAMs. Visualization of the confidence map is shown in \cref{fig:vis_pseudo_label}. We can find that unconfident pixels mainly focus on object boundaries, which is reasonable because boundaries tend to be semantically murky regions.

\begin{table}
  \centering
  \begin{tabular}{lcc}
    \toprule
    Model  & Cross Entropy & CGL \\
    \midrule
    VOC  & 70.6    & 71.1   \\
    VOC$^\ddagger$      & 73.3  & 73.8      \\
    COCO    & 45.1 & 45.4       \\
    \bottomrule
  \end{tabular}
  \caption{Ablation study of Confidence-Guided Loss(CGL).$^\ddagger$ denotes using MS COCO pre-trained model.}
  \label{tab:cgl ablation}
\end{table}

\textbf{Effect of Synonym Fusion.} In \cref{tab:synonym fusion ablation}, we compare performance on some classes with/without synonyms. The result can be improved a lot by applying synonyms, especially for category \textit{``person''}, which we use \textit{``person with clothes, people, human''} to replace.

\begin{table}
  \centering
  \scalebox{0.9}{
  \begin{tabular}{ccccc}
    \toprule
    Category  &bird & chair & person & tvmonitor\\
    \midrule
    Original name & 62.9 & 40.7 & 43.6 & 37.0 \\
    Synonym fusion & 63.9 & 44.1 & 51.6 & 40.3 \\
    \bottomrule
  \end{tabular}}
  \caption{Ablation study of synonym fusion on PASCAL VOC 2012 train set. The results above are based on the initial CAMs and not refined by CAA.}
  \vspace{-4mm}
  \label{tab:synonym fusion ablation}
\end{table}

\section{Conclusion}
This paper explores the potential of CLIP to localize different categories with image-level labels 
and proposes a simple yet effective framework, CLIP-ES, for WSSS. We present several improvement strategies for each stage to obtain high-quality CAMs and reduce the training cost. The novel framework is text-driven and can efficiently generate pseudo masks for semantic segmentation without further training. Our framework achieves state-of-the-art performance on PASCAL VOC 2012 and COCO 2014
and is potential to generate segmentation masks for new classes.

\paragraph{Acknowledgments} This work was supported in part by The National Nature Science Foundation of China (Grant Nos: 62273303, U1909203, 61973271, U1909203), in part by S\&T Plan of Zhejiang Province (No. 202218).

{\small
\bibliographystyle{ieee_fullname}
\bibliography{egbib}
}

\newpage

\appendix
\renewcommand\thefigure{S\arabic{figure}}
\renewcommand\thetable{S\arabic{table}}

\setcounter{figure}{0}
\setcounter{table}{0}

\vspace{-8mm}

\begin{strip}
\begin{center}
  \begin{tabular}{l}
  \LARGE \textbf{Appendix} \\
  \end{tabular}
\end{center}
\end{strip}

\raggedend

\begin{figure}[t]
  \centering
   \includegraphics[width=0.95\linewidth]{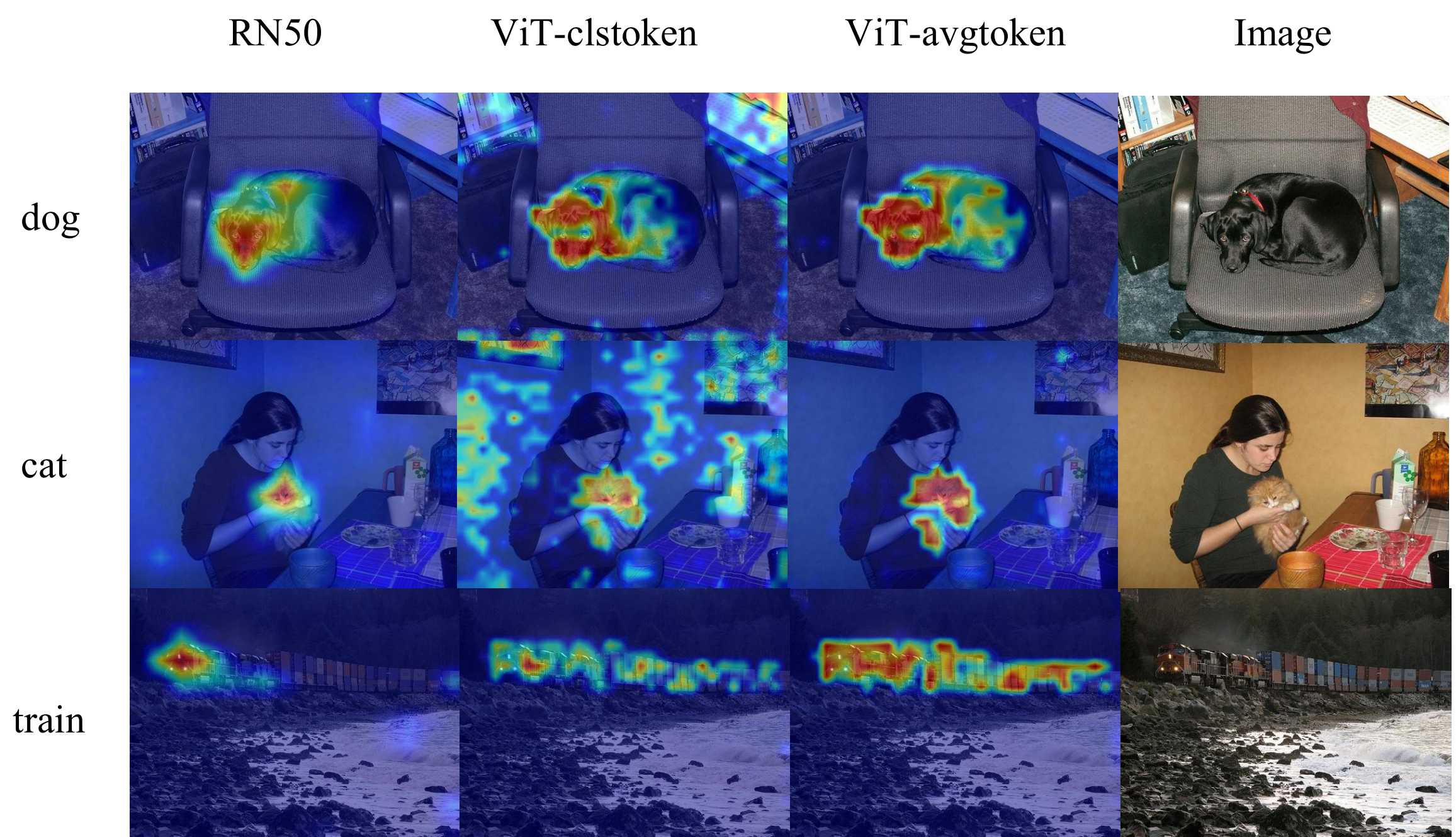}

   \caption{Qualitative comparisons between CNN and ViT architecture as well as clstoken and avgtoken for WSSS task.}
   \label{fig:cnn_vit_cls}
\end{figure}

\section{More Analysis about GradCAM-CLIP}
Pretrained CLIP models include two architectures, \ie, ResNet-based and ViT-based. It is noteworthy that Grad-CAM is not only applicable to CNN-based architecture but also work on vision transformer. In our experiments, we find that the ResNet-based model suffers from the discriminative part domain problem heavily.
In contrast, CAMs generated by ViT tend to cover more parts of objects. The qualitative and quantitative results can be found in \cref{fig:cnn_vit_cls} and \cref{tab:cnn_vit_cls}, respectively. We adopt CLIP-pretrained ViT-B-16 in all our experiments.

Besides, ViT\cite{dosovitskiy2020vit} tends to use an extra class token to get classification logits and compute the loss. An alternative is to perform average pooling on remaining tokens. The classification performances of the two methods tend to be similar in previous works. However, when applying Grad-CAM to CLIP, we find that CAMs generated by these two methods are somewhat different. The latter method can localize objects more completely and accurately, as is shown in \cref{fig:cnn_vit_cls}. We suppose that the classification task is image-level, yet localization is pixel-level or region-level. The clstoken contains semantic information of the whole image and focuses on the patches that contribute more to it, while 
the average value of remaining tokens could treat each token equally. The latter is more suitable for dense prediction tasks, especially for the multi-label setting. Results in \cref{tab:cnn_vit_cls} demonstrate the superiority of the average pooling token for the WSSS task. Furthermore, the \textit{sharpness} of avgtoken is significantly smaller than clstoken. It implies that avgtoken can attend to more classes rather than make one class prominent. The results verify the rationality of our proposed metric as well.

\begin{table}
  \centering
  \begin{tabular}{cccc}
    \toprule
    Model  & Initial & CAA refined & Shaprness \\
    \midrule
    RN50   & 38.2       & -  &  0.019 \\
    ViT-clstoken      & 43.8       & 62.4  &  0.021 \\
    ViT-avgtoken     & 58.6       & 70.8  &  0.004 \\
    \bottomrule
  \end{tabular}
  \caption{Quantitative comparisons between CNN and ViT architecture as well as clstoken and avgtoken for WSSS task.}
  \label{tab:cnn_vit_cls}
\end{table}

\begin{table}
  \centering
  \begin{tabular}{cccc}
    \toprule
    Category  & Sentence-level & Feature-level & CAM-level \\
    \midrule
    bird      & 76.7 & 76.7 & 76.6 \\
    chair     & 48.4 & 48.4 & 47.7 \\
    person    & 63.2 & 63.8 & 65.8 \\
    tvminotor & 57.2 & 57.2 & 53.9 \\
    all classes  & 70.8 & 70.8 & 70.6 \\
    \bottomrule
  \end{tabular}
  \caption{Comparison of different synonym fusion strategies on PASCAL VOC 2012 train set.}
  \label{tab:different synonym fusion}
  \vspace{-2mm}
\end{table}

\section{Comparisons of Different Synonym Fusion Strategies}
We can perform synonym fusion in different stages. Without loss of generality, we divide it into three types: 1) sentence-level (before inputting into text-encoder), 2) feature-level (after text-encoder), 3) CAM-level (after CAM generation). We perform synonym fusion on 4 categories and compare the three strategies in \cref{tab:different synonym fusion}. The results remain similar and merely varied slightly among these approaches for each category as well as all categories. Since the last two methods require multiple encode processes for each synonym, we adopt the time-efficient sentence-level fusion strategy in our experiments.

\section{Hyper-parameter Selection for $\lambda$}
\begin{figure}[t]
  \centering
   \includegraphics[width=0.8\linewidth]{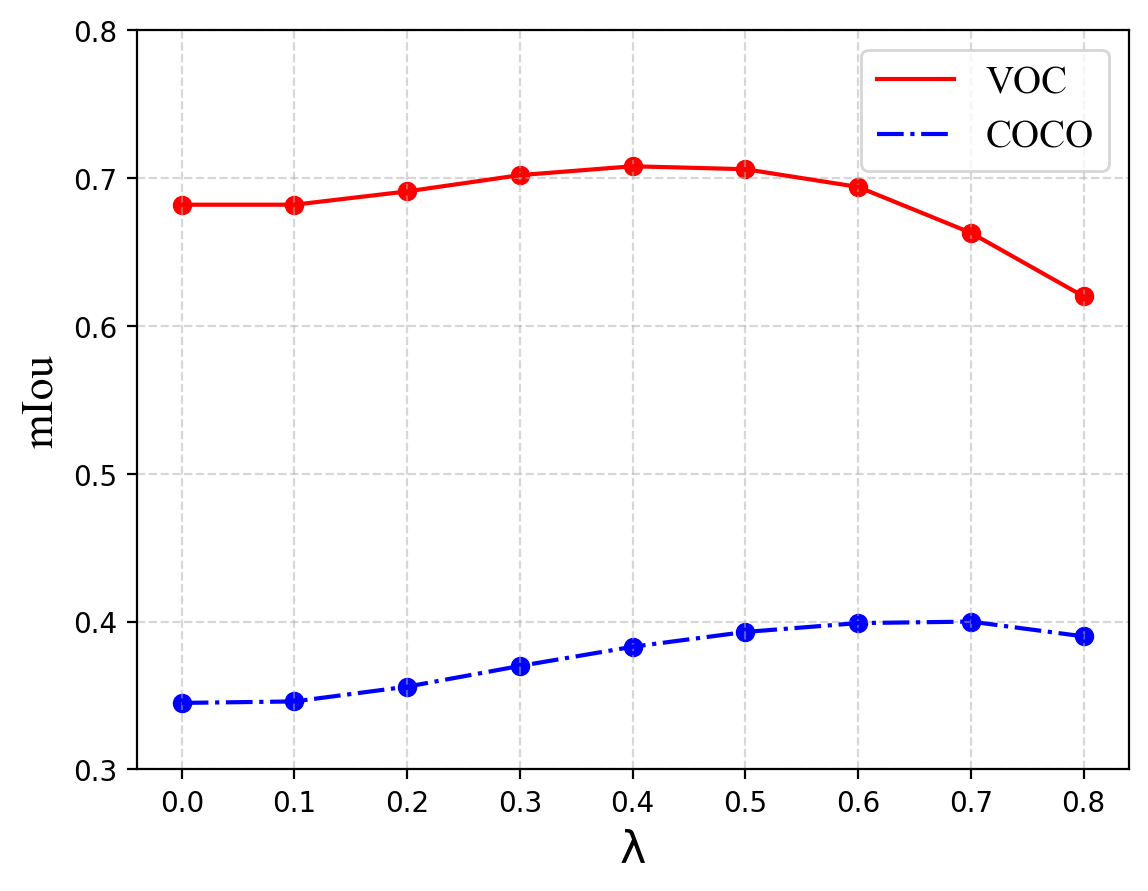}

   \caption{Effect of $\lambda$ for the quality of generated CAMs on PASCAL VOC 2012 and part of COCO 2014 train set.}
   \label{fig:lambda_ablation}
\end{figure}

In CAA module, we generate a class-aware mask for MHSA in the transformer. A parameter $\lambda$ is used to binarize the CAM and generate some bounding boxes. In this part, we investigate the effect of $\lambda$ on PASCAL VOC 2012 and COCO 2014 train set. Since the amount of COCO train set is tremendous, we only select the first 2000 images for research. We vary the threshold from 0 to 0.8 with an interval of 0.1. The results in \cref{fig:lambda_ablation} indicate that the best threshold varies on different datasets. We suppose that COCO is more complex and contains more objects in an image than PASCAL VOC on average. Therefore, a stricter threshold is required to identify regions belonging to the target class. In our experiment, we set $\lambda$ to 0.4 and 0.7 for VOC and COCO, respectively. 

\section{Hyper-parameter Selection for $\mu$ in CGL}
In the experiments, we found most pixels are confident enough after dense CRF postprocessing~\cite{CRF}. We calculate the confidence distribution on VOC (VOC's original ignored percentage is about 5.4\%). Results in \cref{tab:conf_dist} indicate that only a small minority of pixels (mainly near object boundaries) have confidence lower than 0.95, and $\mu$ doesn't affect the segmentation performance remarkably. Therefore, we set $\mu$ to 0.95 in our experiments.

\begin{table}
  \centering
  \scalebox{0.94}{
  \begin{tabular}{cccc}
    \toprule
    Confidence  & [0.5, 0.8] & [0.8, 0.95] & [0.95, 1.0]  \\
    \midrule
    Frequency(\%)   & 0.78 & 1.43 & 97.75  \\
    \bottomrule
    \midrule
    $\mu$  & 0.7 & 0.8 & 0.95  \\
    \midrule
    mIoU  & 73.7 &  73.6 & 73.8 \\
    \bottomrule
  \end{tabular}}
  \vspace{-2mm}
  \caption{The distribution of confidence and mIoU of final segmentation with different $\mu$ on VOC 12.}
  \label{tab:conf_dist}
\end{table}

\section{Training Details of DeepLabV2}
For VOC, images are randomly scaled to [0.5, 0.75, 1.0, 1.25, 1.5] and cropped to 321x321. The batch size is set to 10, and iteration is 20k as default. For COCO, we use strong augment following~\cite{Lee2022AMN}. Images are randomly scaled to [0.5, 0.75, 1.0, 1.25, 1.5, 1.75, 2.0] and 481x481 are cropped. The batch size and the number of training iterations are set to 5 and 100k, respectively. The initial learning rate is 2e-4 for imagenet-pretrained model and 2.5e-5 for COCO-pretrained model, with the polynomial learning rate decay $lr_{iter} = lr_{init}(1 - \frac{iter}{maxiter})^\gamma$, where $\gamma = 0.9$. We set $\mu=0.95$ to ignore unconfident pseudo labels. Balanced cross-entropy loss is adopted for COCO training as in \cite{Lee2021advcam,Lee2022AMN}. For testing, we adopt a multi-scale strategy and dense CRF to post-process with default hyper-parameters in~\cite{chen2017deeplab}.

\section{Detailed Setting of Time and Memory Efficiency}
We compare our proposed framework with classical AdvCAM~\cite{Lee2021advcam}, another language-supervised work CLIMS~\cite{Xie_2022_CLIMS} and ViT-based work MCTFormer~\cite{xu2022mctformer} in term of time and memory. All the experiments are conducted on a TITAN RTX GPU with 24 GB memory. We use their open-source code and follow the default procedure. When applying dense CRF, 20 num-workers are adopted for multiprocessing. The maximum memory occurs during the affinity network training stage, which is about 18GB for both
PSA~\cite{Ahn2018PSA} and IRN~\cite{Ahn2019IRN}. With only 2GB memory, our training-free method could generate pseudo masks for PASCAL VOC 2012 train aug set (with 10582 images) within 1 hour. Note that adopting multiple GPUs or multiprocessing can further speed up this process.

\section{Background Set}
We define 25 class-related background categories for VOC, including \{\textit{ground, land, grass, tree, building, wall, sky, lake, water, river, sea, railway, railroad, keyboard, helmet, cloud, house, mountain, ocean, road, rock, street, valley, bridge, sign}\}. For COCO, we simply remove \{\textit{sign, keyboard}\} since these categories have been defined in COCO categories.

\section{More Qualitative Results}
\begin{figure*}[h]
\vspace{2mm}
  \centering
   \includegraphics[width=18cm, height=19cm]{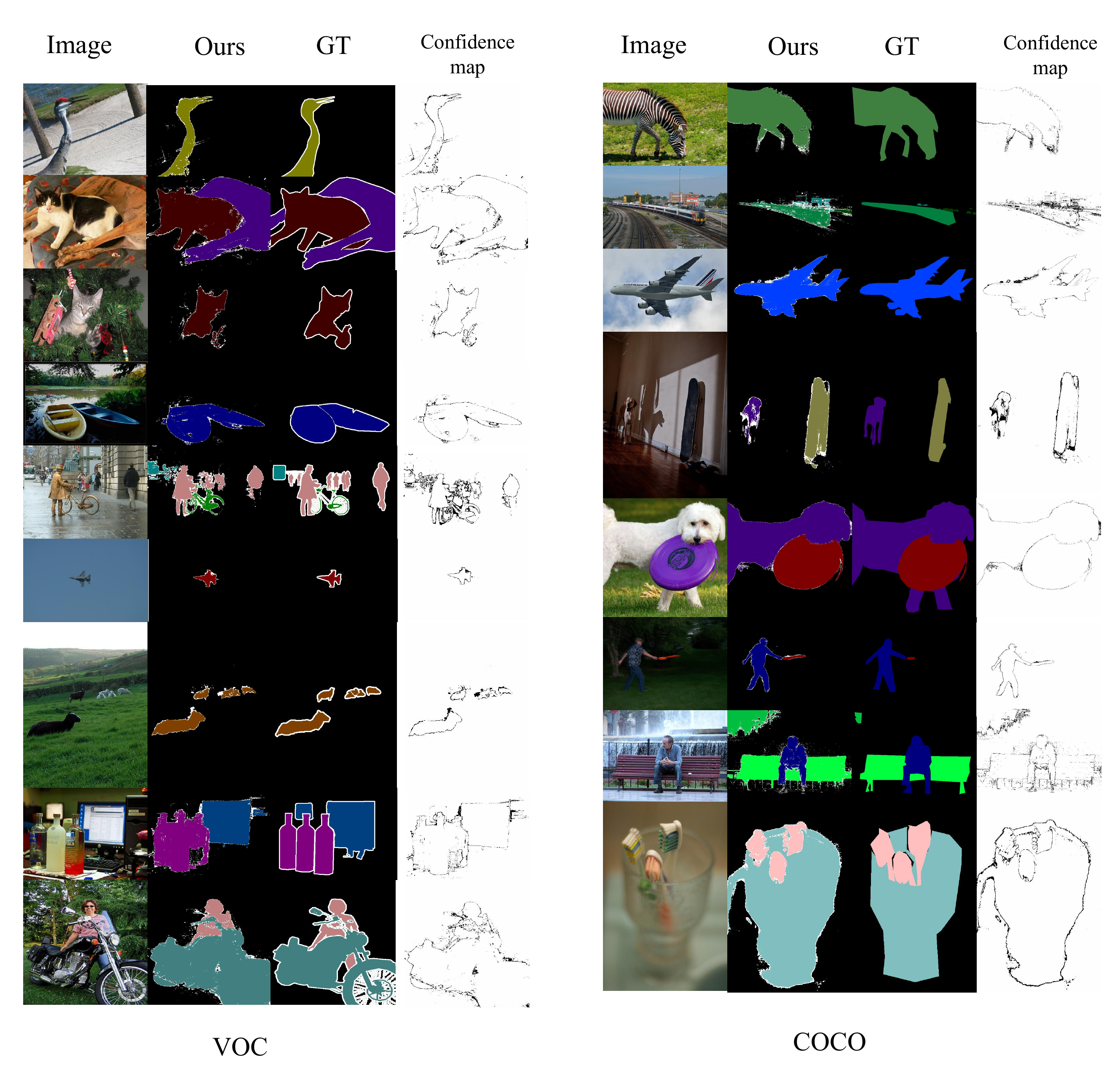}

   \caption{More visualizations on PASCAL VOC 2012 and MS COCO 2014 datasets.
   }
   \label{fig:more_vis}
   \vspace{2mm}
\end{figure*}

In \cref{fig:more_vis}, we provide more qualitative results of our generated pseudo labels and corresponding confidence maps on PASCAL VOC 2012 and MS COCO 2014 datasets. We can observe that our proposed framework produces satisfactory segmentation results. It is effective in both simple and complex scenes.

\end{document}